%% file: 0.main.tex
\pdfoutput=1
\documentclass[11pt]{article}

\usepackage[final]{acl}
\usepackage{amsmath,amssymb,amsfonts}
\usepackage{algorithmic}
\usepackage{float}
\usepackage{amsthm}
\usepackage{times}
\usepackage{latexsym}
\usepackage{overpic}

\usepackage[T1]{fontenc}

\usepackage[utf8]{inputenc}

\usepackage{microtype}

\usepackage{inconsolata}

\usepackage{graphicx}
\usepackage{setspace}

\usepackage{hyperref}

\usepackage{times}
\usepackage[inline]{asymptote}
\usepackage{amssymb}
\usepackage{multirow}
\usepackage{adjustbox} 
\usepackage{svg}
\usepackage{tablefootnote}
\usepackage{listings}
\usepackage{xcolor}
\usepackage{makecell}
\usepackage{multicol}
\usepackage{flushend}
\usepackage{booktabs}
\usepackage{threeparttable}
\usepackage{url}
\usepackage{pgfplots}
\usepackage{pgfplotstable}
\pgfplotsset{compat=1.17}
\usepackage{subcaption}
\usepackage{enumitem}
\usepackage{caption}
\usepackage{threeparttable}
\usepackage{placeins}
\usepackage{amssymb}
\usepackage{pifont}
\usepackage{algorithm2e}
\usepackage[defaultcolor=red,final]{changes}
\usepackage{verbatim}
\usepackage{xkeyval}
\lstdefinestyle{PythonStyle}{
    language=Python,
    basicstyle=\ttfamily\large,
    keywordstyle=\color{blue},
    commentstyle=\color{gray},
    stringstyle=\color{red},
    numberstyle=\tiny\color{gray},
    stepnumber=1,
    numbersep=5pt,
    breaklines=true,
    breakatwhitespace=false,
    tabsize=4,
    showstringspaces=false
}
\usepackage{tikz}
\usepackage{cleveref}
\crefname{section}{Appendix}{Appendices}
\hypersetup{
    colorlinks=true,        
    linkcolor=blue,         
}

\newcommand{\ie}{\textit{i.e.}}

\newcommand{\etc}{\textit{etc.}}


\title{InfiFusion: A Unified Framework for Enhanced Cross-Model Reasoning via LLM Fusion}

\author{
Zhaoyi Yan\textsuperscript{1},
Yiming Zhang\textsuperscript{2},
Baoyi He\textsuperscript{3},
Yuhao Fu\textsuperscript{5},
Qi Zhou\textsuperscript{4},
Zhijie Sang\textsuperscript{1},
Chunlin Ji\textsuperscript{5},\\
\textbf{
Shengyu Zhang\textsuperscript{3},
Fei Wu\textsuperscript{3},
Hongxia Yang\textsuperscript{1,2}*
}\\
\textsuperscript{1} Reallm Labs,
\textsuperscript{2} The Hong Kong Polytechnic University,
\textsuperscript{3} Zhejiang University,\\
\textsuperscript{4} Harbin Institute of Technology,
\textsuperscript{5} Independent \\
\textsuperscript{*}\,Corresponding authors. \quad
\texttt{hongxia.yang@polyu.edu.hk}
}

\begin{document}
\maketitle

\input{0.6.Abstract}

\input{1.Intro}

\input{2.Related}

\input{3.Method}
\input{4.Experiments}

\bibliography{custom}
\appendix

\input{Appendix.Notion}
\input{Appendix.others}

\input{Appendix.case}

\end{document}

%% file: 0.6.abstract.tex
\begin{abstract}
We introduce InfiFusion, an efficient training pipeline designed to integrate multiple domain-specialized Large Language Models (LLMs) into a single pivot model, effectively harnessing the strengths of each source model.
Traditional fusion methods either merge model parameters directly or rely on knowledge distillation with rigid assumptions, limiting their flexibility and efficiency.
InfiFusion overcomes these limitations by enhancing Universal Logit Distillation (ULD) with Top-K selection and Logits Standardization. Top-K selection reduces noise by prioritizing the most informative logits, while logits standardization ensures better alignment between models by normalizing logits across diverse models.
We propose two fusion strategies: Pairwise Fusion (InfiFusion$_p$), where each source model’s knowledge is distilled individually into the pivot model followed by merging and Unified Fusion (InfiFusion$_u$), where knowledge from all source models is distilled simultaneously into the pivot model.
InfiFusion outperforms the state-of-the-art models, such as Qwen-2.5-14B-Instruct and Phi-4, across 11 widely applied benchmarks covering reasoning, coding, mathematics, and instruction-following tasks.
Notably, InfiFusion achieves this superior performance while significantly reduces computational costs, completing full training with only 160 H800 GPU hours compared to the millions typically required for traditional LLM training.
\end{abstract}

%% file: 1.Intro.tex
\section{Introduction}
\noindent Large Language Models (LLMs) have demonstrated remarkable success across various tasks such as general reasoning, mathematics, programming, and scientific applications \cite{zhu2024deepseek, qwen2025qwen25technicalreport}. However, no single model excels in all domains, and training a comprehensive model from scratch requires significant computational resources. An alternative approach is to fuse multiple domain-specialized models into a single pivot model, combining their strengths without requiring retraining a monolithic model.

\noindent Traditional LLM merging techniques~\cite{TA, yadav2024ties, wan2024fusechat} rely on direct integration of model parameters, but these methods presuppose uniform architectures across models and often fail to capture the strengths of diverse specialized models.
LLM fusion typically transfers expertise from source models to the pivot model via knowledge distillation, supporting models with varying architectures and vocabularies.
However, the performance is strongly dependent on vocabulary and token alignment, limiting their applicability.
The Universal Logit Distillation (ULD) framework \cite{boizard2025crosstokenizerdistillationuniversallogit} addresses this by using the 1-Wasserstein distance to align token distributions across models with different vocabularies. Despite its potential, ULD faces challenges, like zero-padding noise and inefficiencies in large-scale models.

\begin{figure}[!t]
  \centering
  \scriptsize
  \begin{subfigure}[b]{.48\columnwidth}
    \includegraphics[width=\linewidth]{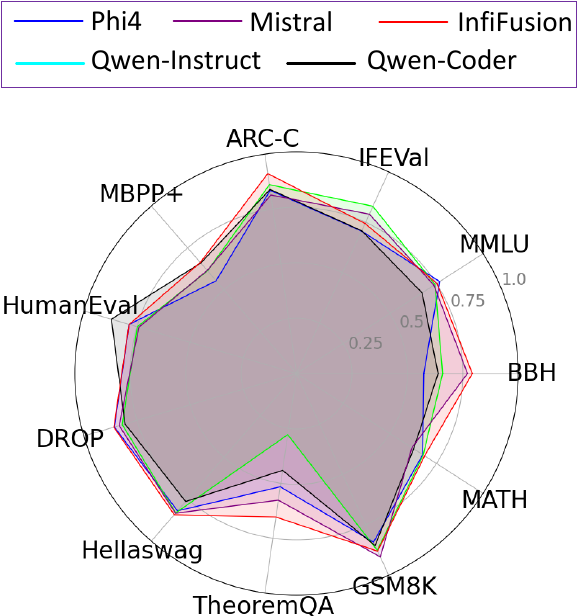}
    \caption{Pivot Model: Phi-4}
    \label{fig:math_levels1}
  \end{subfigure}
  \begin{subfigure}[b]{.48\columnwidth}
    \includegraphics[width=\linewidth]{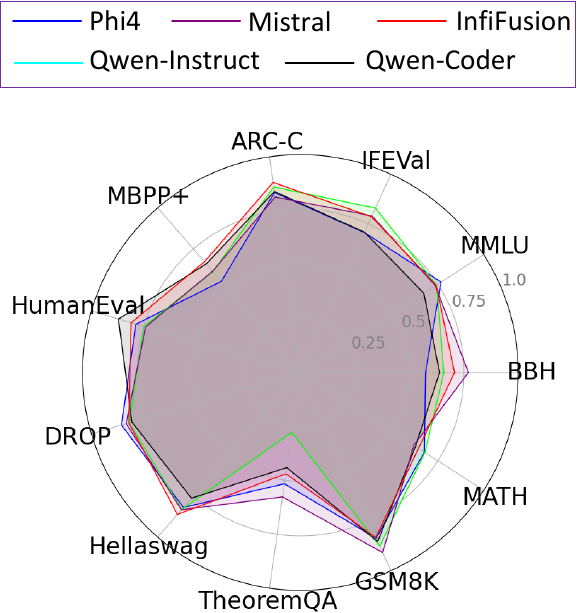}
    \caption{Pivot Model: Qwen-14B}
    \label{fig:math_levels2}
  \end{subfigure}
  \par\vspace{0.5cm} 
  \begin{subfigure}[b]{1\columnwidth}
    \includegraphics[width=\linewidth]{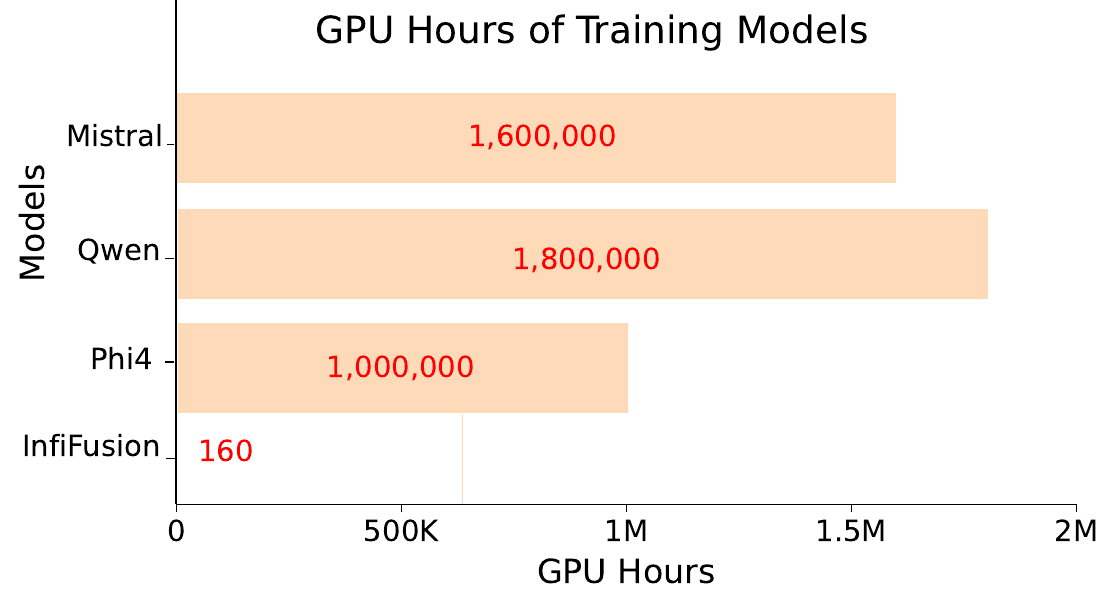}
    \caption{Training GPU hours for different models.}
    \label{fig:math_levels3}
  \end{subfigure}
  \caption{Performance of InfiFusion on the pivot models Phi-4 and Qwen2.5-14B-Instruct. In all cases, InfiFusion significantly outperforms the pivot model in terms of average score. Notably, InfiFusion requires only approximately $1/10000$ of the GPU hours compared to LLMs, such as Qwen2.5-14B-Instruct, Phi-4, \etc.}
  \label{figs:fusion_qwen_phi}
  \vspace{-0.4cm}
\end{figure}

\noindent In this paper, we introduce InfiFusion, a novel fusion framework that enhances ULD by incorporating Top-K selection and logits standardization~\cite{sun2024logit}.
Top-K selection filters out irrelevant logits, reducing noise and computational cost, while Logits Standardization ensures that the pivot model aligns the relative relationships between logits rather than their magnitudes.
These innovations improve the efficiency and robustness of knowledge transfer, making it possible to integrate specialized models across different domains.
We propose two fusion strategies: Pairwise Fusion (\ie, InfiFusion$_{p}$), where each source model’s knowledge is distilled separately into the pivot model, and Unified Fusion (\ie, InfiFusion$_{u}$), which aggregates knowledge from all source models simultaneously.
As shown in Fig.~\ref{figs:fusion_qwen_phi}, InfiFusion outperforms state-of-the-art models over 11 tasks while significantly reducing computational costs.
Unlike traditional models, which demand millions of GPU hours, InfiFusion achieves superior performance with just 160 H800 GPU hours, making it an exceptionally efficient approach.

\noindent Our contributions are summarized as follows:
\begin{itemize}[topsep=0pt,itemsep=0pt,parsep=0pt,partopsep=0pt]
\item We introduce InfiFusion, a novel framework that improves knowledge transfer between domain-specialized models by enhancing Universal Logit Distillation (ULD) with Top-K selection and Logits Standardization.
\item We propose two fusion strategies: Pairwise Fusion and Unified Fusion, each offering distinct advantages in distilling knowledge from multiple source models into a single pivot model.
\item We demonstrate the effectiveness of InfiFusion through extensive experiments, showing that it outperforms state-of-the-art models while significantly reducing computational costs.
\end{itemize}

%% file: 2.Related.tex
\section{Related Work}
\subsection{Model Merging}
Model merging integrates models with the same architecture at the parameter level. Wortsman et al.\cite{wortsman2022model} employed a linear averaging method to merge models that were fine-tuned from the same base model. Matena and Raffel~\cite{matena2022mergingmodelsfisherweightedaveraging} introduced Fisher-based model merging, using Laplace approximation and the Fisher information matrix to enhance parameter merging beyond simple averaging. Ilharco et al.~\cite{ilharco2022editing} used task vectors to represent the differences among source LLMs and applied task arithmetic on these task vectors to achieve model merging. TIES-Merging~\cite{yadav2024ties} builds upon this by trimming redundant values and resolving sign conflicts. DARE~\cite{yu2023language} facilitates the sparsification of task vectors, eliminating over 90\% of the delta parameters while maintaining performance. The SCE ~\cite{wan2024fusechat} method is computationally efficient and memory-conserving, merging parameter matrices using a three-step process. Channel merging~\cite{zhang2024channelmergingpreservingspecialization} clusters and merges channel parameters into several groups offline, significantly reducing parameter conflicts.

\subsection{Knowledge Distillation}
Knowledge Distillation, introduced by~\cite{kd2015}, enables transferring knowledge from large models to a smaller one while preserving performance. Early approaches focused on aligning the student model’s final layer outputs with the teacher’s, known as logits distillation~\cite{distilbert19, wellreadstu19}.
Feature distillation~\cite{patient2019} further improves this by utilizing intermediate teacher model features. Distillation can be classified as on-policy~\cite{austin2021program, distillm24, policy24, minillm24} or off-policy~\cite{distilbert19, patient2019, wellreadstu19, zhang2024dualspaceknowledgedistillationlarge, boizard2024towards} depending on whether the data is generated during training.
While knowledge distillation is widely studied, most research focuses on single-teacher paradigms~\cite{policy24, minillm24, llava24, compact24}, with fewer works exploring multi-teacher combinations, which face challenges like vocabulary mismatch, model parameter differences, and task discrepancies~\cite{babyllama23, ensemble23, wan2024fusellm, wan2024fusechat}.

\subsection{Model Fusion}
Model fusion combines multiple language models to enhance performance and generalization by integrating their predictions or intermediate representations. Unlike traditional merging methods, fusion does not require parallel deployment or models with identical architectures. 
Fusion techniques typically rely on logits fusion or distribution matrix fusion, achieved through manually designed functions~\cite{guo2020online, pham2023collaborative, wan2024knowledge, wan2024fusechat}. For instance, MiniLogit~\cite{guo2020online} proposes logits fusion for multi-teacher collaborative learning in image classification tasks.
FuseLLM~\cite{wan2024knowledge} introduces a method for fusing diverse LLMs by leveraging their generative distributions, while FuseChat~\cite{wan2024fusechat} proposes a two-stage framework: first, applying pairwise knowledge fusion to create target LLMs with identical structure and size, and then merging these models within the parameter space.
In contrast, our approach offers a highly efficient fusion method based on knowledge distillation.
%
%

%% file: 3.Method.tex
\section{Methods}
\subsection{Preliminary}
We consider a pivot language model (LLM) $M_o$ and $S$ source LLMs $\{M_1, \ldots, M_S\}$, each specializing in distinct domains or tasks. Our objective is to distill the combined strengths of these source models into a single, integrated pivot model $M_o$. Given a dataset $\mathcal{D}$ of instruction--response pairs, we apply knowledge distillation to guide $M_o$ in acquiring complementary expertise from the source models. This process involves defining suitable fusion (distillation) objectives and designing fusion pipelines that integrate diverse knowledge without compromising the overall model quality.

\begin{figure*}[!t]  
\centering 
\begin{overpic}[scale=0.9]{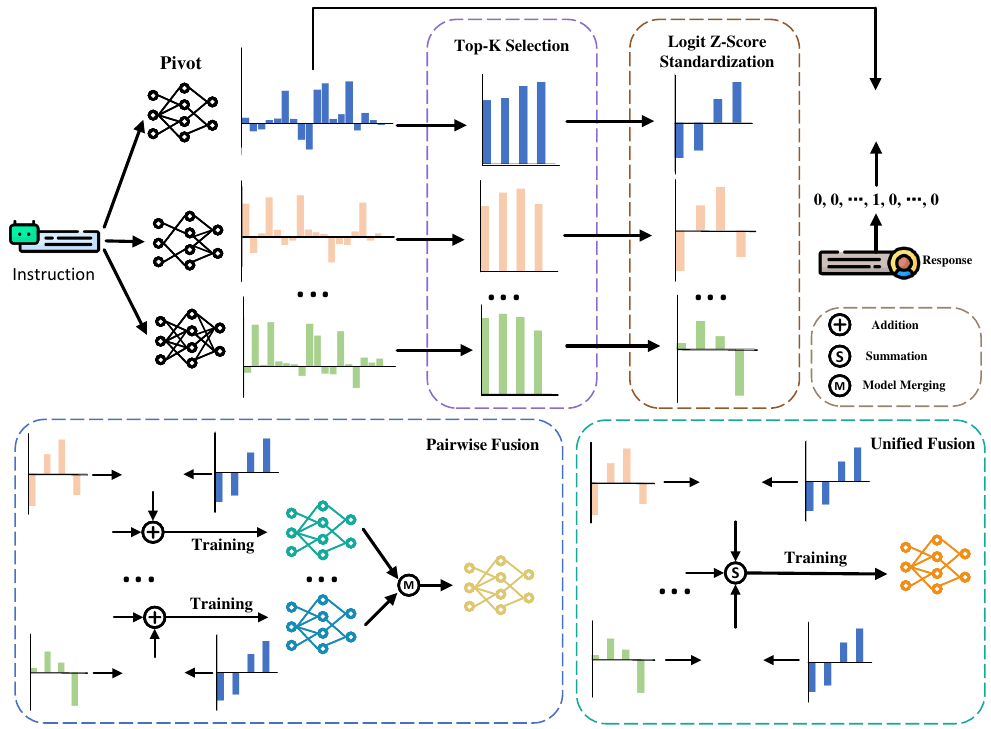}

\put(5, 52){\scalebox{1.0}{$x_i$}}
\put(16.5, 57){\scalebox{1.0}{$M_o$}}
\put(16.5, 44){\scalebox{1.0}{$M_1$}}
\put(16.5, 33){\scalebox{1.0}{$M_S$}}

\put(1.6, 20){\scalebox{0.7}{$(1-\lambda)\mathcal{L}_{\text{SFT}}$}}

\put(1.6, 11){\scalebox{0.7}{$(1-\lambda)\mathcal{L}_{\text{SFT}}$}}
\put(12.4, 5){\scalebox{0.7}{$\lambda\mathcal{L}_{\text{pair},S}$}}
\put(12.4, 25.3){\scalebox{0.7}{$\lambda\mathcal{L}_{\text{pair},1}$}}

\put(30, 25){\scalebox{0.8}{$M_{o,1}$}}
\put(30, 5){\scalebox{0.8}{$M_{o,S}$}}
\put(47.5, 8){\scalebox{0.8}{$M_{o,\text{pair}}$}}

\put(37, 58){\scalebox{0.8}{$z_{o}$}}
\put(37, 47){\scalebox{0.8}{$z_{1}$}}
\put(37, 35){\scalebox{0.8}{$z_{S}$}}

\put(56, 58){\scalebox{0.8}{$l_{o}$}}
\put(56, 47){\scalebox{0.8}{$l_{1}$}}
\put(56, 35){\scalebox{0.8}{$l_{S}$}}

\put(76, 58){\scalebox{0.8}{$\hat{l}_{S}$}}
\put(76, 47){\scalebox{0.8}{$\hat{l}_{S}$}}
\put(76, 35){\scalebox{0.8}{$\hat{l}_{S}$}}

\put(87, 44.0){\scalebox{1.0}{$y_i$}}
\put(86, 62){\scalebox{0.8}{$\mathcal{L}_{\text{SFT}}$}}
\put(58.5, 15.5){\scalebox{0.7}{$(1-\lambda)\mathcal{L}_{\text{SFT}}$}}
\put(71.3, 25){\scalebox{0.7}{$\lambda\mathcal{L}_{\text{pair},1}$}}
\put(71.3, 6.5){\scalebox{0.7}{$\lambda\mathcal{L}_{\text{pair},S}$}}
\put(92, 11){\scalebox{0.8}{$M_{o,\text{uf}}$}}

\end{overpic}
\caption{Illustration of the InfiFusion framework, incorporating Top-K selection and logits standardization. Two fusion strategies are proposed: Pairwise Fusion, where each source model’s knowledge is distilled separately into the pivot model, and Unified Fusion, which aggregates knowledge from all source models simultaneously.}
    \label{fig:pdf_example}
\end{figure*}
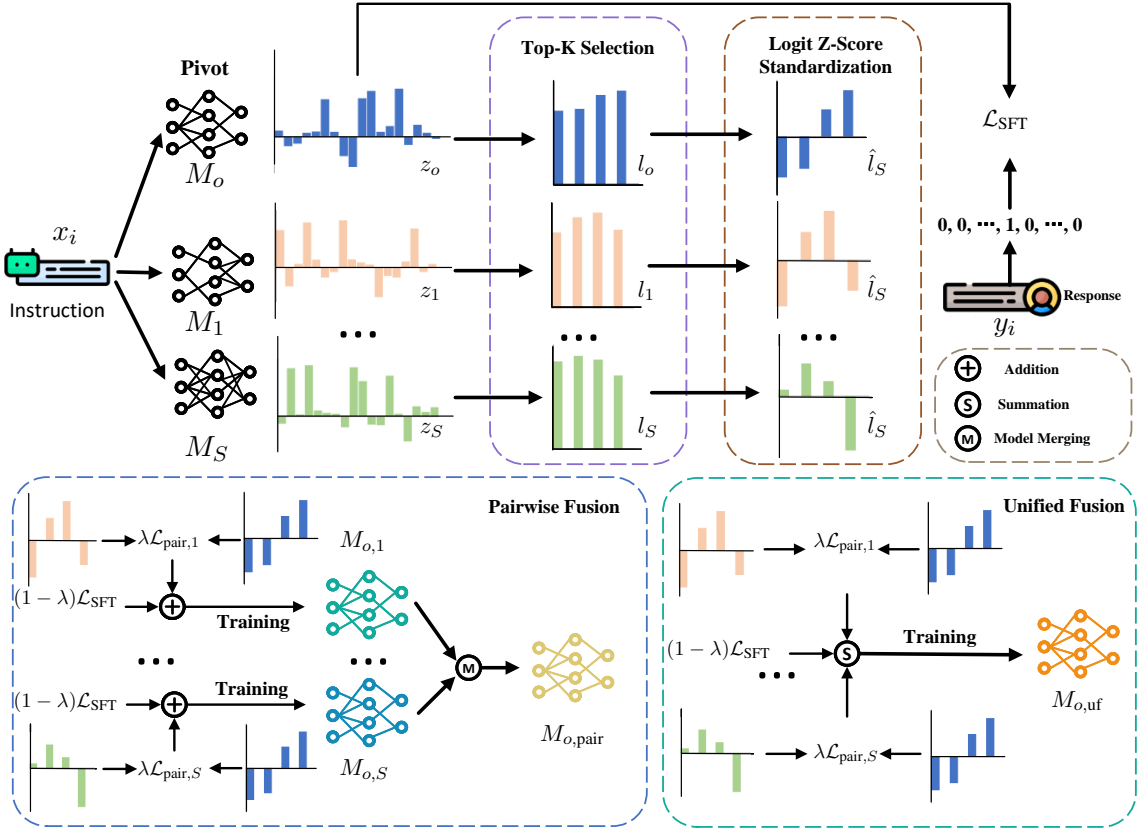

\subsubsection{Supervised Fine-Tuning (SFT)}
Supervised fine-tuning (SFT) aligns the pivot model $M_o$ with instruction-response pairs in the fine-tuning dataset $\mathcal{D}$. For each instruction $x_i$ and its corresponding response $y_i$ of length $T$, the SFT objective is:
\begin{equation}
\label{eq:sft}
\mathcal{L}_{\text{SFT}} 
= 
-\,\mathbb{E}_{(x_i, y_i) \sim \mathcal{D}} \Bigl[ \sum_{t=1}^{T} \log p_{\theta_o}\bigl(y_{i,t} \mid y_{i,<t}, x_i\bigr) \Bigr],
\end{equation}
where $y_{i,<t}$ denotes the tokens preceding $y_{i,t}$ and $p_{\theta_{o}}(\cdot)$ is the model-predicted token probability under $M_o$.

\subsubsection{Optimal Transport Loss}
OT provides a powerful way to compare two discrete distributions by solving a “mass transport” problem. Formally, let 
$p$ and $q$ be two discrete distributions over vocabularies of size 
$V_o$ and $ V_s$, respectively.. We define:
\begin{equation}
\mathrm{OT}(p, q)
= 
\min_{\gamma \in \Pi(p,q)}
\sum_{m=1}^{V_o} \sum_{n=1}^{V_s} \gamma_{m,n} \, C_{m,n}.
\label{eq:ot-def}
\end{equation}
Here, $\Pi(p,q)$ is the set of joint distributions with
marginals $p$ and $q$.
The cost matrix $C_{m,n}$ encodes the cost of ``transporting'' one unit of mass from token $m$ to $n$.
In practice, Sinkhorn-based methods with entropic regularization can be used to approximate~\eqref{eq:ot-def} efficiently.

\subsubsection{Discrete 1-Wasserstein}
An important special case of OT is the 1-Wasserstein distance (also called the Earth Mover’s Distance). In discrete settings, $\mathcal{W}_1(p,q)$ is obtained by choosing a particular cost matrix. For instance, if we set $C_{m,n}=1$ (i.e., uniform cost whenever $m\ne n$), the 1-Wasserstein distance has a closed-form solution:
\begin{equation}
    \mathcal{W}_{1}(p,q)=\sum^{V}_{i=1}|\mbox{sort}(p)_i-\mbox{sort}(q)_i|
\end{equation}
Let $V=\max \bigl(V_o, V_s\bigr)$ be the effective vocabulary size, we augment either the student or teacher vocabulary size through distribution padding (with 0 value), ensuring equal support size for both. 
where $sort(p)$ rearranges the probabilities in 
$p$ in descending order. Intuitively, under uniform cost, the optimal transport simply pairs “largest probability mass” with ``largest probability mass,'' the second-largest with the second-largest, and so on. This makes $\mathcal{W}_1$ particularly simple and effective for aligning distributions whose labels (tokens) may not match one-to-one.

\subsubsection{Universal Logit Distillation Loss}
\label{sec:uld}

To effectively transfer knowledge from multiple source models to the pivot model $M_o$, we adopt \emph{Universal Logit Distillation} (ULD) loss~\cite{uld24}, which aligns probability distributions across different vocabularies using a structured distance metric. Given a source model $M_s$ with vocabulary $\mathcal{V}_s$ and the pivot model $M_o$ with vocabulary $\mathcal{V}_o$, their predicted probability distributions at time step $t$ are denoted as $P_{\theta_o}(\cdot \mid \Tilde{x}_{<t})$ and $P_{\theta_s}(\cdot \mid \hat{x}_{<t})$, respectively. Since these distributions may not be directly comparable due to vocabulary mismatches, ULD loss mitigates this issue by minimizing the 1-Wasserstein distance:
\begin{equation}
\label{eq:uld}
\mathcal{L}_{\text{ULD}} =
\sum_{t=1}^{T} 
\mathcal{W}_1\big(
    P_{\theta_o}(\cdot \mid \Tilde{x}_{<t}), 
    P_{\theta_s}(\cdot \mid \hat{x}_{<t})
  \big).
\end{equation}
Here, $\mathcal{W}_1$ measures the effort required to transform one probability distribution into another by optimally aligning their probability masses. Unlike KL-divergence, which is sensitive to label mismatches, ULD loss provides a more robust way to distill knowledge across different vocabularies while preserving the semantic structure of the predictions.

\subsection{Top-K Selection and Logits Standardization}

To improve the distillation process, we introduce two strategies: Top-K selection and logits standardization. These techniques aim to enhance knowledge transfer while addressing temperature sharing and vocabulary mismatches between the pivot model $M_o$ and the source models ${M_1, \ldots, M_S}$.

\paragraph{Top-K Selection}
We focus on the top-$K$ logits for both the pivot model $M_o$ and each source model $M_s$, instead of using the entire logit distribution. This is based on the observation that most logits in instruction-based models correspond to low-probability categories, contributing little to the distillation process.
By selecting the top-$K$ logits, we reduce noise from irrelevant categories, improving both knowledge transfer quality and computational efficiency. For high-performance instruction models, such as Phi-4, a small value of $K$ (e.g., $K=10$) captures nearly all the probability mass, enhancing distillation while reducing the computational burden of comparing all logits.

\paragraph{Logits Standardization}  
To address challenges in temperature sharing, we apply logits standardization to the logits before distillation. Traditional methods assume a shared temperature between the models, which can cause performance issues when the logits' ranges differ. 
By normalizing the logits, we decouple the magnitude of the logits from the distillation process, allowing the pivot model to focus on their relative order and structure. The logits for both models are normalized as follows:
\[
\hat{l}_{o, t} = \frac{l_{o, t} - \mu_{o}}{\sigma_{o}}, \quad \hat{l}_{s, t} = \frac{l_{s, t} - \mu_{s}}{\sigma_{s}},
\]
where $\mu_o$, $\sigma_o$, $\mu_s$, and $\sigma_s$ are the mean and standard deviation of the top-K logits for the pivot and source models, respectively.
This ensures that the pivot model aligns effectively with the source models, focusing on the relative relationships between logits rather than their magnitudes. Combined with Top-K selection, this preprocessing step ensures more robust and efficient distillation, even when logit distributions differ.

\subsection{Fusion Strategies}

\subsubsection{Pairwise Fusion}
To integrate knowledge from multiple source models $\{M_1, \dots, M_S\}$ into the pivot model $M_o$, we adopt a pairwise fusion strategy, where each source model’s knowledge is distilled into $M_o$ separately.
This fusion is performed by minimizing the Universal Logit Distillation (ULD) loss, incorporating both Top-K selection and logit normalization to enhance the alignment process.

\noindent For each source model $M_s$, we first apply Top-K selection to the original logits $z_s$, retaining the most informative logits $l_s$, i.e., $l_s = \text{TopK}(z_s, K)$. We then normalize the logits using logits standardization.
The resulting loss is computed as:
\begin{equation}
\mathcal{L}_{\text{pair}, s}= \mathbb{E}_{(x_i, y_i) \sim \mathcal{D}} \Big[ \sum_{t=1}^{T} \Big| \hat{l}_{\text{o}, t} - \hat{l}_{\text{s}, t} \Big| \Big] ,
\end{equation}
where $\hat{l}_{\text{o}, t}$ and $\hat{l}_{\text{s}, t}$ are the normalized logits for the student and teacher models, respectively, at time step $t$.
By minimizing this loss, we encourage the student model to align its predictions with the teacher’s on the most probable tokens, while reducing the impact of irrelevant logits. Let $M_{o,s}$ be the optimized pivot model, its parameters are $\theta_{o,s}$ are obtained via:

\begin{equation}
\label{eqn:pair_k_sft}
    \theta_{o,s} = \arg\min_{\theta_o}\lambda \mathcal{L}_{\text{pair}, s} + (1-\lambda )\mathcal{L}_{\text{SFT}}
\end{equation}
$\lambda$ is the hyper-parameter balancing the losses.
By looping $s$ from $1$ to $S$, which indicates we perform $S$ times of training, we obtain $M_{o,1}, \cdots M_{o,S}$.
Given $\{M_{o,s}\}_{s=1}^{S}$ share the same architecture, following ~\cite{wan2024fusechat}, they can be merged via model merging methods and obtain the final model $M_{o,\text{pair}}$.
\begin{equation}
    M_{o, \text{pair}} = \text{Merge}(M_{o,1}, \cdots, M_{o,S})
\end{equation}


\subsubsection{Unified Fusion}
To aggregate knowledge from multiple teacher models, we adopt a unified fusion strategy. In this approach, the knowledge from each source model is distilled into the pivot model by computing the pairwise fusion loss for each teacher, summing the losses, and combining them with the supervised fine-tuning (SFT) loss to form the final objective function. For each source model $M_s$, the pairwise fusion loss $\mathcal{L}_{\text{pair}, s}$ is computed using Top-K selection and logits standardization. The total loss is then given by:
\begin{equation}
\mathcal{L}_{\text{uf}} = \lambda\sum_{s=1}^{S} \mathcal{L}_{\text{pair}, s} + (1 - \lambda) \mathcal{L}_{\text{SFT}},
\end{equation}
where $\lambda$ controls the balance between the distillation and supervised fine-tuning losses. The parameters of the pivot model $M_o$ are optimized as:
\begin{equation}
\theta_{o, \text{uf}} = \arg\min_{\theta_o} \lambda \sum_{s=1}^S \mathcal{L}_{\text{pair}, s} + (1 - \lambda) \mathcal{L}_{\text{SFT}}.
\end{equation}
The result of this fusion process is the unified pivot model $M_{o, \text{uf}}$, which benefits from both the labeled data in the fine-tuning dataset and the distilled knowledge from the source models, focusing on the most relevant logits via Top-K selection and logits standardization.

\noindent Pairwise fusion and unified fusion are proposed as two independent approaches for achieving multi-source model fusion. They are based on different assumptions. Specifically, unified fusion defines an explicit global loss and assumes that the optimal weights for this loss will yield the best performance. In contrast, pairwise fusion defines an objective loss implicitly. It aims to achieve local optimization and assumes that the average of these local optima will result in the best overall performance. We provide a detailed theoretical analysis about the relation between pairwise fusion and unified fusion in Appendix~\ref{app:relation}.

%% file: 4.Experiments.tex
\section{Experiments}
\subsection{Settings}
\textbf{Dataset} In our fusion experiments, we create an InfiFusion-Mix dataset, which consists of 180,000 samples used for facilitating knowledge transfer from the source model to the pivot model.
The samples in this dataset are sourced from Infinity-Instruct~\cite{li2024superfiltering}, ScaleQuest-Math~\cite{ding2024unleashing}, and opc-sft-stage2~\cite{huang2024opencoder}. Detailed information about the InfiFusion-Mix dataset can be found in Table~\ref{tab:fusion_data}.
The InfiFusion-Mix dataset is split into train and test sets at an 8:2 ratio, resulting in 144k training and 36k test samples.

\begin{table}[!h]
\centering
\begin{adjustbox}{width=\linewidth}
{
\begin{tabular}{lccc}
\toprule
Types &  \textbf{General Data} & \textbf{Math Data} & \textbf{Code Data}\\
\midrule
Dataset & Infinity-Instruct~\tablefootnote{\tiny https://huggingface.co/datasets/BAAI/Infinity-Instruct} & ScaleQuest-Math~\tablefootnote{\tiny https://huggingface.co/datasets/dyyyyyyyy/ScaleQuest-Math} & opc-sft-stage2~\tablefootnote{\tiny https://huggingface.co/datasets/OpenCoder-LLM/opc-sft-stage2} \\
Original Size & 1.4M & 1.0M & 436K \\
Sample Size & 80K & 50K & 50K \\
\bottomrule
\end{tabular}
}
\end{adjustbox}
\vspace{-0.1in}
\caption{The details of InfiFusion-Mix dataset}
\label{tab:fusion_data}
\end{table}

\noindent\textbf{Training Details} 
We train all models using the C-AdamW optimizer~\cite{liang2024cautiousoptimizersimprovingtraining} with a cosine annealing scheduler for a total of 5 epochs. Early stopping is applied at the 4th epoch to prevent overfitting and reduce GPU costs. The entire training of InfiFusion$_u$ takes approximately 20 hours, utilizing a global batch size of 16 across $8\times80$GB NVIDIA H800 GPUs, with a learning rate of $1 \times 10^{-6}$. By default, we set $\lambda = 0.5$.
For a detailed comparison of GPU hours for different methods, please refer to Table~\ref{tab:gph_h}.

\noindent\textbf{Offline Teacher Loading} Since the source models are static, we can pre-extract their hidden states (before the \textit{lm\_head} layer). For each source model, the hidden states are first compressed and stored on disk, requiring approximately 1.5 TB for InfiFusion-Mix.
Each source model only retains a single fully connected layer (i.e., \textit{lm\_head}), which significantly speeds up the InfiFusion training process, making it comparable to the training of pure SFT models. Moreover, this approach greatly reduces the GPU memory footprint of fusion.

\noindent\textbf{Evaluation} We evaluate our models across multiple domains, using 11 different benchmarks that cover a wide range of fields, including general reasoning, mathematics, coding, and text reasoning. Specifically, we assess the model's mathematical capabilities with benchmarks such as GSM8K~\cite{cobbe2021gsm8k}, Math~\cite{hendrycks2021math}, and TheoremQA~\cite{chen2023theoremqa}. For coding, we use MBPP+~\cite{austin2021program} and HumanEval~\cite{chen2021humaneval}, while general reasoning is evaluated through BBH~\cite{suzgun2022challenging}, ARC-C~\cite{yadav2019quick} and MMLU~\cite{hendrycks2021measuringmassivemultitasklanguage}. We also test instruction-following ability with IFEval~\cite{zhou2023instruction}, and text reasoning performance using DROP~\cite{dua2019drop} and Hellaswag~\cite{zellers2019hellaswag}. 

\noindent\textbf{Pivot Model and Source Models} Before choosing the pivot model, we evaluate the performance of state-of-the-art open-source LLMs across 11 benchmarks in Table~\ref{tab:popolar_models_main_results}. To maintain both high training efficiency and strong performance, we select the powerful instruction-following model Phi-4~\cite{abdin2024phi} as the pivot model, and use Qwen2.5-Coder-14B-Instruct~\cite{qwen2025qwen25technicalreport}, Qwen2.5-14B-Instruct~\cite{qwen2025qwen25technicalreport}, and Mistral-Small-24B-Instruct-2501\footnote{\tiny \url{https://huggingface.co/mistralai/Mistral-Small-24B-Instruct-2501}} as the source models.
By fusion, the strengths of the source models are aggregated and integrated into the pivot model.
\begin{table*}[htbp]
\setlength{\tabcolsep}{2pt} 
\scriptsize
\centering
\setlength{\extrarowheight}{1pt}
\resizebox{1\hsize}{!}{
\begin{tabular}{l| c c c| c c | c c c | c | c c | c | c | c } 
\toprule
\multirow{2}{*}{\textbf{Models}} & \multicolumn{3}{c|}{\textbf{Math}} & \multicolumn{2}{c|}{\textbf{Coding}} & \multicolumn{3}{c|}{\textbf{General Reasoning}} & \multicolumn{1}{c|}{\textbf{Instruct Following}} & \multicolumn{2}{c|}{\textbf{Text Reasoning}} & \multirow{2}{*}{\textbf{Avg}} & \multirow{2}{*}{\textbf{Params}} & \textbf{GPU}\\
{} & GSM8K & MATH & TheoremQA & MBPP+ & HumanEval & BBH & ARC-C & MMLU & IFEval & DROP & Hellaswag & {} & {}  & {\textbf{hours}}\\
\midrule
\multicolumn{14}{c}{\textbf{Pivot Model}} \\
\midrule
Phi-4 & 86.88 & 74.10 & 36.12 & 64.29 & 82.93 & 65.90 & 86.78 & \textbf{81.38} & 76.50 & 88.49 & 85.57 & 75.08 & 14B & \textasciitilde\text{1.0M} \\
\midrule
\multicolumn{14}{c}{\textbf{Source Model}} \\
\midrule
Qwen2.5-14B-Coder& 88.25 & 70.24 & 30.88 & 72.75 & \textbf{89.63} & 71.07 & 87.12 & 73.84 & 76.62 & 84.49 & 81.08 & 75.09 & 14B & \textasciitilde\text{1.8M} \\
Qwen2.5-14B-Instruct& 90.14 & 74.32 & 19.50 & 69.04 & 79.88 & 72.70 & 88.81 & 79.38 & \textbf{86.33} & 85.50 & 86.55 & 75.65 & 14B & \textasciitilde\text{1.8M} \\
Mistral-small-24B & \textbf{92.72} & 69.48 & 40.38 & 67.22 & 79.27 & 81.67 & 85.08 & 78.92 & 83.21 & 86.68 & 86.71 & 77.39 & 24B & \textasciitilde\text{1.6M} \\
\midrule
\multicolumn{14}{c}{\textbf{SFT}}  \\
\midrule
SFT (Phi-4) & 90.75 & 71.96 & 37.88 & 71.00 & 78.05 & 83.03 & 92.88 & 80.28 & 77.22 & 87.69 & 86.79 & 77.96 & 14B & 145 \\
\midrule
\multicolumn{14}{c}{\textbf{Fusion Competitors} } \\
\midrule
SeqKD* & 89.23 & 73.06 & 28.62 & 69.80 & 81.10 & 81.03 & 92.88 & 79.97 & 82.13 & 87.40 & 75.34 & 76.41 & 14B & 580 \\
MiniLogit* & 90.33 & 73.34 & 41.49 & 69.22 & 82.41 & 81.36 & 91.19 & 80.15 & 78.41 & 87.56 & 86.99 & 78.40 & 14B & 220 \\
FuseLLM* & 89.95 & 72.76 & 41.28 & 69.44 & 81.34 & 82.17 & 91.63 & 79.74 & 78.94 & 86.97 & 86.86 & 78.28 & 14B & 225 \\
FuseChat* & 89.86 & 74.21 & 44.06 & 70.18 & 82.19 & 81.77 & 91.08 & 79.66 & 78.09 & 87.94 & 86.67 & 78.70 & 14B & 650 \\

\midrule
\multicolumn{14}{c}{\textbf{Pairwise Fusion}} \\
\midrule
InfiFusion$_{p}$\& TA & 90.98 & 74.70 & \textbf{46.50} & 72.20 & 82.93 & \textbf{83.62} & 92.20 & 80.34 & 79.86 & \textbf{88.73} & 87.52  & \textbf{79.96} & 14B & 450 \\
InfiFusion$_{p}$\& TIES & 90.22 & 74.32 & 42.12 & 72.75 & 85.98 & 83.19 & 91.19 & 80.48 & 79.14 & 88.53 & \textbf{87.73} & 79.60 & 14B & 450 \\
InfiFusion$_{p}$\& SCE & 90.22 & 74.40 & 42.00 & 70.60 & 85.37 & 83.24 & 91.19 & 80.37 & 80.34 & 88.60 & 87.72 & 79.46 & 14B & 450 \\
\midrule
\multicolumn{14}{c}{\textbf{Unified Fusion}} \\
\midrule
InfiFusion$_{u}$ & 90.45 & \textbf{74.92} & 45.75 & \textbf{73.02} & 82.93 & 83.44 &\textbf{ 92.88} & 80.18 & 79.50 & 88.71 & 87.38 & \textbf{79.92} & 14B & 160 \\
\bottomrule
\end{tabular}}
\vspace{-0.1in}
\caption{Performance of models on state-of-the-art benchmarks, with GPU hours estimated on the GPU H800. InfiFusion$_u$ achieves nearly identical performance ($\textbf{79.92}$ vs. $\textbf{79.96}$) compared to InfiFusion${_p}$\&TA, while utilizing only approximately $\mathbf{35\%}$ of the GPU hours (theoretically approximated as $1/S$, and here $S=3$).
InfiFusion$_u$ achieves \textbf{79.92} (vs. \textbf{75.08} for Phi-4), while requiring only about \textbf{0.016\%} of the GPU hours.
TA, TIES, and SCE represent the merging method proposed in ~\cite{TA}, ~\cite{yadav2024ties} and ~\cite{wan2024fusechat}, respectively.
\textbf{*} denotes methods re-implemented and trained on our dataset, pivot, and source models.}
\label{tab:main_results}
\end{table*}
\vspace{-1em}
\subsection{Main Results}
\subsubsection{Pairwise and Unified Fusion}
\noindent\textbf{Pairwise Fusion} In pairwise fusion, the pivot and source models are grouped into several model pairs. Each pair is then trained using the pairwise fusion loss, resulting in $S$ fused models.
These $S$ fused models share the same architecture as the pivot model but have distinct weights. They are subsequently merged using commonly used methods, such as Task Arithmetic~\cite{TA}, Ties-Merging~\cite{yadav2024ties}, and SCE~\cite{wan2024fusechat}.
As shown in Table~\ref{tab:main_results}, the pairwise fused model outperforms the pivot model, Phi-4, across 10 benchmarks. Specifically, the InfiFusion$_p$\&TA model achieves an average score improvement of 4.88 points. Additionally, InfiFusion$_p$ significantly surpasses the source models, with average score improvements of 4.87, 4.31, and 2.57 for Qwen2.5-14B-Coder, Qwen2.5-14B-Instruct, and Mistral-small-24B, respectively.
We observe that TIES and SCE exhibit moderate performance drops compared to TA. We attribute this to the presence of a hyperparameter, $\tau$, in both methods, which demands precise tuning.
In this study, we use the default value to emphasize the robustness of the InfiFusion approach.
While more precise tuning of this hyperparameter could potentially enhance performance, such an investigation lies beyond the scope of this paper.

\noindent\textbf{Unified Fusion}
In unified fusion, all source models are fused with the pivot model simultaneously.
However, this approach presents two key challenges: token misalignment and conflicts arising from the diversity of source models. To address these, we introduce a ULD loss and a Top-K selection mechanism. The ULD loss provides an effective way to measure the distance between distributions of different sizes, mitigating issues caused by token misalignment.
Meanwhile, the Top-K selection reduces noise from small logits and alleviates conflicts stemming from the varying styles of the source models. For more details, see Table~\ref{tab:abl_zscore_topk}.

\noindent As shown in Table~\ref{tab:main_results}, unified fusion, which leverages the strengths of the source models, significantly outperforms the pivot model by 4.84 points.
Similarly, when compared to the source models, InfiFusion$_u$ surpasses them, with average score improvements of 4.83, 4.27, and 2.53 for Qwen2.5-14B-Coder, Qwen2.5-14B-Instruct, and Mistral-small-24B, respectively.
InfiFusion$_u$ delivers performance nearly identical to InfiFusion$_p$\&TA, with only a 0.04-point difference on average, while consuming only 35\% of the GPU hours required by InfiFusion$_p$\&TA.
A detailed comparison of GPU hours between pairwise and unified fusion is presented in Table~\ref{tab:gph_h}.

\subsubsection{Comparison with State-of-the-art Methods}
We evaluate five methods: SFT, SeqKD~\cite{kim2016seqkd}, MiniLogit~\cite{guo2020online}, FuseLLM~\cite{wan2024fusellm}, and FuseChat~\cite{wan2024fusechat}.
SFT involves directly fine-tuning the Phi-4 model on our InfiFusion-Mix dataset with only the SFT loss. The performance improves from 75.08 to 77.96, demonstrating the effectiveness of InfiFusion-Mix.
With the incorporation of the fusion loss, performance improves further to 79.92 for InfiFusion$_u$ and 79.96 for InfiFusion$_p$, highlighting the efficacy of our fusion approach.
SeqKD generates data using all the source models, where each model generates the same volume of InfiFusion-Mix data, thereby expanding the training dataset. However, despite the addition of $S$ (here $S=3$) times more generated data, SeqKD results in worse performance than SFT. This suggests that simply adding more generated data may lead to confusion for the pivot model.
MiniLogit and FuseLLM both utilize unified fusion. MiniLogit, a commonly used logit fusion method for image classification tasks, and FuseLLM, which applies a simple MinCE distribution fusion for LLMs, both yield lower performance than InfiFusion. This is attributed to the superiority of ULD with Top-K selection and logits standardization compared to the KL divergence loss used by these methods.
FuseChat adopts pairwise fusion, achieving a moderate higher average score than FuseLLM, but at the cost of requiring nearly $S$ (here $S=3$) times more GPU resources for training.

\subsection{Ablation Study}

\noindent\textbf{Logits Standardization and Top-K Selection.}
To evaluate the impact of logits standardization and Top-K selection, we compare models trained with and without these techniques.
As shown in Table~\ref{tab:abl_zscore_topk}, applying only Top-K selection results in a notable performance improvement, from 79.02 to 79.66. In contrast, logits standardization alone yields a smaller gain. When both techniques are combined, InfiFusion$_u$ achieves further performance improvements, reaching 79.92.
These results suggest that focusing on the most informative logits can be effective.

\begin{table}[!h]
\centering
\resizebox{0.95\hsize}{!}{
\begin{tabular}{lcccc}
\toprule
\textbf{Method} & Top-K Selection & Logits Standardization  & Avg \\
\midrule
InfiFusion$_{u}$ & {} & {} & 79.02 \\
InfiFusion$_{u}$ & \checkmark & {} & 79.66 \\
InfiFusion$_{u}$ & {} & \checkmark & 79.29 \\
InfiFusion$_{u}$ & \checkmark & \checkmark & \textbf{79.92} \\
\bottomrule
\end{tabular}
}
\vspace{-0.1in}
\caption{Impact of Logits Standardization and Top-K Selection on InfiFusion$_u$}
\label{tab:abl_zscore_topk}
\end{table}

\noindent\textbf{Number of Source Models.}
Next, we explore the effect of increasing the number of source models in the fusion process.
Specifically, we investigate two configurations: (1) fusing the pivot model with a single source model, and (2) fusing the pivot model with multiple source models simultaneously using the pairwise/unified fusion strategy. Table~\ref{tab:gph_h} presents the results of this ablation study.

\begin{table}[!hbtp]
\centering
\begin{adjustbox}{width=\linewidth}
{
\begin{tabular}{lccccc}
\toprule
\multicolumn{6}{c}{\textbf{Pivot model: Phi-4}}\\
\midrule
Models &  Qwen-Coder & Qwen-Instruct & Mistral & Avg. & GPU Hours\\
\midrule
Phi-4 SFT & - & - & - & 77.96 & 145 \\
\midrule
\multirow{4}{*}{InfiFusion$_p$} & \checkmark & \checkmark &  & 79.56 & 300 \\
& \checkmark &  & \checkmark & 79.40 & 300\\
&  & \checkmark & \checkmark & 79.39 & 300\\
& \checkmark & \checkmark & \checkmark & \textbf{79.96} & 450\\
\midrule
\multirow{4}{*}{InfiFusion$_u$} & \checkmark &  & &  78.70 & 150\\
&  & \checkmark & &  78.65 & 150\\
&  &  & \checkmark &  79.61 & 150\\
& \checkmark & \checkmark & &  79.54 & 155\\
& \checkmark &  & \checkmark & 79.45 & 155\\
&  & \checkmark & \checkmark & 79.31 & 155 \\
& \checkmark & \checkmark & \checkmark & \textbf{79.92} & 160\\
\bottomrule
\end{tabular}
}
\end{adjustbox}
\vspace{-0.1in}
\caption{Performance of InfiFusion with varying numbers of source models}
\label{tab:gph_h}
\end{table}

\noindent It is evident that when fusing a single source model, the average score improves from 77.96 to 78.70, 78.65, and 79.61, for Qwen2.5-Coder-14B, Qwen2.5-14B-Instruct and Mistral-small-24B, respectively.
As more source models are fused, both pairwise and unified fusion strategies yield substantial performance improvements.
Notably, InfiFusion$_u$ exhibits performance close to that of InfiFusion$_p$, while requiring approximately $1/S$ of the training GPU hours.

\noindent\textbf{Value of $K$.}
We also investigate the effect of varying the value of $K$ in Top-K selection. 
From Table~\ref{tab:abl_k_value}, we observe that when $K$ is too small (i.e., $K\!<\!10$), the knowledge transfer quality decreases due to too few logits being considered during fusion.
Conversely, when $K$ is too large (i.e., $K\!>\!10$), noise is introduced, which degrades performance. 
Consequently, we select $K\!=\!10$ as it yields the best performance. 
Interestingly, this aligns with our observation that for high-performance instruction models like Phi-4, a small value of $K$ (e.g., $K\!=\!10$) captures nearly all of the probability mass.

\begin{table}[!hbtp]
\centering
\begin{adjustbox}{width=0.9\linewidth}
{
\begin{tabular}{lcccccc}
\toprule
{$K$} & 5 & 10 & 15 & 20 & 25 \\
\midrule
Avg.     & 79.54    & \textbf{79.92}   & 79.46    & 79.41    & 79.39 \\
\bottomrule
\end{tabular}
}
\end{adjustbox}
\vspace{-0.1in}
\caption{Impact of $K$ on InfiFusion$_{u}$ performance}
\label{tab:abl_k_value}
\end{table}


\noindent\textbf{Dataset Scaling.}
We conduct an ablation study to assess the impact of data scaling on InfiFusion by varying the size of the training dataset. Specifically, we train models with datasets ranging from 50k to 144k instruction-response pairs. 
As shown in Table~\ref{tab:data_scale}, increasing the training data size consistently improves InfiFusion's performance. This emphasizes the value of dataset scaling and indicates that our method benefits from larger datasets, leading to enhanced performance.

\begin{table}[!hbtp]
\centering
\begin{adjustbox}{width=0.85\linewidth}
{
\begin{tabular}{lcccc}
\toprule
{Dataset Scale} & 50,000 & 100,000 & 144,000 \\
\midrule
Avg.      & 79.24    & 79.49    & \textbf{79.92}    \\
\bottomrule
\end{tabular}
}
\end{adjustbox}
\vspace{-0.1in}
\caption{InfiFusion$_{u}$ performance on varied data scales}
\label{tab:data_scale}
\end{table}

\section{Conclusion}
We present InfiFusion, an innovative framework that seamlessly integrates domain-specialized LLMs into a unified pivot model, enhancing Universal Logit Distillation (ULD) through Top-K selection and logits standardization. InfiFusion effectively distills knowledge from diverse source models, overcoming vocabulary mismatches and computational inefficiencies. Two fusion strategies, Pairwise Fusion (InfiFusion$_p$) and Unified Fusion (InfiFusion$_u$), offer flexibility in merging expertise.
Experiments reveal that InfiFusion surpasses state-of-the-art models, such as Qwen-2.5-14B-Instruct and Phi-4, across 11 benchmark tasks, achieving superior performance with only 160 H800 GPU hours. InfiFusion stands as an efficient and scalable solution for high-performance LLM deployment.

\newpage

\section{Limitations}
While our unified fusion approach offers a straightforward and promising framework, it does not consistently outperform the pairwise fusion strategy. Additionally, the current method is limited to large language models (LLMs), and the fusion of models similar to GPT-O1 has not yet been explored.
Another limitation is that the source models should not lag too far behind the pivot model in terms of performance, which restricts the selection of viable source models.
In future work, we aim to relax this constraint, allowing for more flexible model selection. We also plan to extend our fusion approach to include both traditional instruct LLMs and GPT-O1-like models, thereby broadening the scope of model capabilities.

%% file: Appendix.Notion.tex
\appendix
\onecolumn
\section*{Appendix A: The Relationship between Unified and Pairwise Fusion}
\label{app:relation}
In this session, we analyze the relationship between unified fusion and pairwise fusion. First, we will generalize the loss function for unified fusion.
\begin{equation}
\mathcal{L}_{\text{uf}} 
= \lambda_0\mathcal{L}_{\text{SFT}} + \sum_{s=1}^{S} \lambda_s\mathcal{L}_{\text{pair},s}
\end{equation}
where $\sum\lambda_s = 1$, $\mathcal{L}_{\text{SFT}}$ and $\mathcal{L}_{\text{pair},s}$ is defined in equation \ref{eq:sft} and \ref{eq:uld}, respectively. Therefore, to find the optimal $\theta$ for the Unified is to solve the optimization problem:
\begin{align}
\label{eq:theta_aio}
\theta_{o,\text{uf}} & = \arg\min_{\theta_o} \lambda_0\mathcal{L}_{\text{SFT}} + \sum_{s=1}^{S} \lambda_s\mathcal{L}_{\text{pair},s}
\end{align}
In the scenario of the pairwise fusion, we can obtain $\{\theta_{o,s}\}_{s=1}^{S}$ fused models, where we have
\begin{equation}
\label{eq:theta_k}
\theta_{o,s} = \arg\min_{\theta_o} \alpha_s\beta_s\mathcal{L}_{\text{SFT}} + \alpha_s(1-\beta_s)\mathcal{L}_{\text{pair},s}.
\end{equation}
where $\alpha_s$ is a scaling factor and $\sum \alpha_s=1$. Let 
\begin{equation}
\mathcal{L}_{s} 
=  \alpha_s\beta_s\mathcal{L}_{\text{SFT}} + \alpha_s(1-\beta_s)\mathcal{L}_{\text{pair},s}.
\end{equation}
Note that when $\lambda_0 = \sum_{s=1}^{S}\alpha_s\beta_s$, $(1-\beta_s) = \lambda_s$, we have
\begin{equation}
 \mathcal{L}_{\text{uf}}(\theta) =  \sum_{s=1}^{S}\mathcal{L}_{s}(\theta).
\end{equation}
The optimization problem \ref{eq:theta_aio} can be rewritten as
\begin{equation}
    \theta_{o,\text{uf}} = \arg\min_{\theta_o} \sum_{s=1}^{S}\mathcal{L}_{s}(\theta_o),
\end{equation}
, and the equation \ref{eq:theta_k} for $s=1,2...,S$, can be written as:
\begin{align}
    \{\theta_{o,s}\}_{s=1}^{S} &= \arg\min_{\theta_{o}^1,...,\theta_{o}^S}\sum_{s=1}^{S}\mathcal{L}_{s}(\theta_{o}^s).
\end{align}
The final fused model $\theta_{o, \text{ps}}$ is derived via weight averaging
\begin{align}
    \theta_{o,\text{ps}} & = \sum_{s=1}^{S}\alpha_s\theta_{o,s} 
\end{align}
In unified fusion, the loss $\mathcal{L}$ can be explicitly written in a function of $\mathcal{L}_i$
\begin{equation}
\mathcal{L}(\mathcal{L}_{1},\mathcal{L}_{2},...,\mathcal{L}_{S}) = \sum_{s=1}^{S}\mathcal{L}_{s}
\end{equation}
,whereas for the pairwise fusion, $\mathcal{L}$ is not explicitly defined. However, we can assume that the $\mathcal{L}$ exists such that
when
\begin{equation}
    \frac{\partial \mathcal{L}_s}{\partial\theta}|_{\theta = \theta_{o,s}}=0,
\end{equation}
we have
\begin{equation}
    \frac{\partial \mathcal{L}}{\partial\theta}|_{\theta = \sum_{s=1}^{S}\alpha_s\theta_{o,s}}=0.
\end{equation}
Ideally, unified fusion can infinitely approximate the global optimum of its defined objective function, but it highly depends on the effectiveness of the loss function. Therefore, a well-designed loss, which is favorable for network optimization, mitigates the conflicts between source models and reflects the real world evaluation becomes essential. By comparison, pairwise fusion looks like an experimental tool, as no theory can support the assumption that the average of local optima can lead to the high performance. The future work should be centered around improving the merging methods, instead of simply averaging but more complexity combinations of the local optima.

%% file: Appendix.others.tex
\section*{Appendix B: Model Selection}
The guiding philosophy of InfiFusion is to elevate the capabilities of smaller LLMs through the strategic fusion and integration of models, thereby enabling the efficient attainment of high-performance, compact LLMs.
Accordingly, we focus our evaluation on state-of-the-art models with fewer than 30B parameters.
For the purpose of efficient training on a single node equipped with $8\times80$GB H800 GPUs, we impose an additional constraint, limiting the pivot model to fewer than 20B parameters.
To achieve optimal performance, we select Phi-4 as the pivot model and include Qwen2.5-14B-Instruct, Qwen2.5-Coder-14B-Instruct, and Mistral-Small-24B-Instruct-2501. These models have demonstrated performance either on par with or exceeding that of the pivot model.

\begin{table*}[htbp]
\setlength{\tabcolsep}{2pt} 
\scriptsize
\centering
\setlength{\extrarowheight}{1pt}
\resizebox{1\hsize}{!}{
\begin{tabular}{l| c c c| c c | c c c | c | c c| c | c } 
\toprule
\multirow{2}{*}{\textbf{Models}} & \multicolumn{3}{c|}{\textbf{Math}} & \multicolumn{2}{c|}{\textbf{Coding}} & \multicolumn{3}{c|}{\textbf{General Reasoning}} & \multicolumn{1}{c|}{\textbf{Instruct Following}} & \multicolumn{2}{c|}{\textbf{Text Reasoning}} & \multirow{2}{*}{\textbf{Params}} & \multirow{2}{*}{\textbf{Avg}} \\
{} & GSM8K & MATH & TheoremQA & MBPP+ & HumanEval & BBH & ARC-C & MMLU & IFEval & DROP & Hellaswag & {} & {} \\
\midrule
Opencoder-8B-Instruct & 31.54 & 10.92 & 14.50 & 63.80 & 81.10 & 46.08 & 49.83 & 42.24 & 54.08 & 39.69 & 43.20 & 8B & 43.36 \\
Llama3.1-8B-Instruct & 79.78 & 49.04 & 32.38 & 59.40 & 68.29 & 67.78 & 81.69 & 69.50 & 85.13 & 82.34 & 73.84 & 8B & 68.11 \\
phi-4 & 86.88 & 74.10 & 36.12 & 64.29 & 82.93 & 65.90 & 86.78 & 81.38 & 76.50 & 88.49 & 85.57 & 14B & 75.08  \\
DeepSeek-Coder-V2-Lite-Instruct & 85.06 & 60.20 & 19.88 & 72.40 & 75.00 & 62.22 & 71.53 & 62.62 & 62.83 & 74.63 & 68.01 & 14B & 64.94\\
Qwen2.5-14B-Instruct & 90.14 & 74.32 & 19.50 & 69.04 & 79.88 & 72.70 & 88.81 & 79.38 & 86.33 & 85.50 & 86.55 & 14B & 75.65\\
Qwen2.5-Coder-14B-Instruct & 88.25 & 70.24 & 30.88 & 72.75 & 89.63 & 71.07 & 87.12 & 73.84 & 76.62 & 84.49 & 81.08 & 14B & 75.09\\
Falcon3-10B-Instruct & 54.66 & 46.46 & 45.88 & 66.20 & 71.95 & 75.78 & 51.53 & 74.63 & 83.33 & 78.06 & 81.57 & 10B & 66.36\\
Llama-3.1-Tulu-3-8B & 82.26 & 45.56 & 29.25 & 47.00 & 63.41 & 68.73 & 65.76 & 65.81 & 87.41 & 76.64 & 71.74 & 8B & 63.96\\
Qwen-2.5-Coder-7B-Instruct & 81.73 & 62.10 & 22.62 & 72.20 & 87.80 & 60.45 & 84.41 & 66.16 & 69.30 & 77.92 & 76.20 & 7B & 69.17\\
starcoder2-15b-instruct-v0.1 & 24.41 & 13.96 & 18.00 & 62.40 & 65.85 & 18.99 & 44.75 & 49.60 & 46.40 & 50.91 & 24.05 & 15B & 38.12\\
OpenMath2-Llama3.1-8B & 85.37 & 65.68 & 25.25 & 14.00 & 9.15 & 45.33 & 45.08 & 32.29 & 32.01 & 74.42 & 32.06 & 8B & 41.88\\
internlm3-8b-instruct & 85.14 & 72.46 & 18.62 & 57.80 & 80.49 & 55.62 & 77.97 & 73.26 & 86.81 & 83.39 & 89.76 & 8B & 71.03 \\
Mistral-Small-24B-Instruct-2501 & 92.72 & 69.48 & 40.38 & 67.22 & 79.27 & 81.67 & 85.08 & 78.92 & 83.21 & 86.68 & 86.71 & 24B & 77.39\\
\bottomrule
\end{tabular}}
\vspace{-0.1in}
\caption{The performance of models on the state-of-art benchmarks}
\label{tab:popolar_models_main_results}
\end{table*}

%% file: Appendix.case.tex
\section*{Appendix C: Case study}
Table~\ref{case_study} shows three examples where Phi-4 is wrong but InfiFusion$_{u}$ is correct. These three problems come from GSM8K, HumanEval, and Hellaswag, corresponding to math, coding, and text reasoning domains. Phi-4 misunderstood the phrase `two times older than' in the question from GSM8K. For the problem from HumanEval, Phi-4 got the index of the number for the quantity of oranges wrong. Phi-4 also failed to make the correct reasoning in the question from Hellaswag, where Option A, `cuts something on a knife board using an electric tool,' represents an impossible action. These errors were corrected in InfiFusion$_{u}$, indicating that InfiFusion$_{u}$ indeed performs better on more tasks and possesses stronger reasoning capabilities.

\begin{table*}[ht]
\centering

\begin{adjustbox}{width=\linewidth}
\begin{tabular}{lp{6cm}p{10cm}p{8cm}}
\toprule
Models & 
GSM8K prompt: \newline
Brandon's iPhone is four times as old as Ben's iPhone. Ben's iPhone is two times older than Suzy's iPhone. If Suzy's iPhone is 1 year old, how old is Brandon's iPhone? Please reason step by step, and put your final answer within boxed{}.
& 
HumanEval prompt:
\begin{lstlisting}[language=Python]
def fruit_distribution(s, n):
    """
    In this task, you will be given a string that represents a number of apples and oranges 
    that are distributed in a basket of fruit this basket contains 
    apples, oranges, and mango fruits. Given the string that represents the total number of 
    the oranges and apples and an integer that represent the total number of the fruits 
    in the basket return the number of the mango fruits in the basket.
    for example:
    fruit_distribution("5 apples and 6 oranges", 19) -> 19 - 5 - 6 = 8
    fruit_distribution("0 apples and 1 oranges", 3) -> 3 - 0 - 1 = 2
    fruit_distribution("2 apples and 3 oranges", 100) -> 100 - 2 - 3 = 95
    fruit_distribution("100 apples and 1 oranges", 120) -> 120 - 100 - 1 = 19
    """
    # Function implementation here
    pass
\end{lstlisting}
& 
Hellaswag prompt: \newline
A woman is reading a newspaper nearby while he speaks to her and shows her something in his hand. The man sharpens his knives using an electric tool. He Question: Which ending makes the most sense? A. sharpens and cuts something on a knife board using an electric tool. B. then continues cutting his onion. C. travels with his knife under the table and uses it also in his shoe. D. then shows her a spare knife that he has also used on a knife sharpener and drills into a side of the table. You may choose from 'A', 'B', 'C', 'D'. Answer: \\
\midrule
Phi-4 & 
{\small
To solve the problem, we need to determine the ages of the iPhones based on the relationships given.

1. \textbf{Determine Suzy's iPhone age:}
   \begin{itemize}
       \item Suzy's iPhone is given as 1 year old.
   \end{itemize}

2. \textbf{Determine Ben's iPhone age:}
   \begin{itemize}
       \item Ben's iPhone is two times older than Suzy's iPhone.
       \item If Suzy's iPhone is 1 year old, then Ben's iPhone is \( 1 + 2 \times 1 = 3 \) years old.
   \end{itemize}

3. \textbf{Determine Brandon's iPhone age:}
   \begin{itemize}
       \item Brandon's iPhone is four times as old as Ben's iPhone.
       \item Since Ben's iPhone is 3 years old, Brandon's iPhone is \( 4 \times 3 = 12 \) years old.
   \end{itemize}

Thus, Brandon's iPhone is \( \boxed{12} \) years old. }
&
\begin{lstlisting}[language=Python]
def fruit_distribution(s, n):
    # Split the string into words
    words = s.split()
    
    # Extract the number of apples and oranges
    apples = int(words[0])
    oranges = int(words[4])
    
    # Calculate the number of mangoes
    mangoes = n - apples - oranges
    
    return mangoes
\end{lstlisting}
& 
The most logical ending is: A. sharpens and cuts something on a knife board using an electric tool. This option follows the context of the man sharpening his knives and suggests a continuation of his activity related to knife use. \\
InfiFusion$_{u}$ & 
{\small
To determine the age of Brandon's iPhone, we need to follow the relationships given in the problem step by step.

1. \textbf{Determine the age of Ben's iPhone:}
   \begin{itemize}
       \item We know that Suzy's iPhone is 1 year old.
       \item Ben's iPhone is two times older than Suzy's iPhone.
       \item Therefore, the age of Ben's iPhone is:
   \end{itemize}
   \[
   2 \times 1 = 2 \text{ years old}
   \]

2. \textbf{Determine the age of Brandon's iPhone:}
   \begin{itemize}
       \item Brandon's iPhone is four times as old as Ben's iPhone.
       \item Since Ben's iPhone is 2 years old, the age of Brandon's iPhone is:
   \end{itemize}
   \[
   4 \times 2 = 8 \text{ years old}
   \]

Thus, the age of Brandon's iPhone is \( \boxed{8} \).
}
& 
\begin{lstlisting}[language=Python]
def fruit_distribution(s, n):
    # Split the string by spaces
    fruits = s.split()
    # Initialize the count of apples and oranges to 0
    apples = 0
    oranges = 0
    # Iterate through the fruits list
    for i in range(len(fruits)):
        # If the current element is "apples", get the number of apples
        if fruits[i] == "apples":
            apples = int(fruits[i-1])
        # If the current element is "oranges", get the number of oranges
        elif fruits[i] == "oranges":
            oranges = int(fruits[i-1])
    # Calculate the number of mangoes
    mangoes = n - apples - oranges
    # Return the number of mangoes
    return mangoes

\end{lstlisting} 
& 
B. then continues cutting his onion.\\
\bottomrule 

\end{tabular}
\end{adjustbox}
\caption{Case study: Three examples where Phi-4 is wrong but InfiFusion$_{u}$ is correct.}
\label{case_study}
\end{table*}